\documentclass[letterpaper, 11 pt, conference]{ieeeconf}  

\IEEEoverridecommandlockouts                              

\usepackage{graphics} 
\usepackage{epsfig} 
\usepackage{mathptmx} 
\usepackage{times} 
\usepackage{amsmath} 
\usepackage{amssymb}  
\usepackage{graphicx}
\usepackage{mathtools}
\usepackage{algorithmic}
\usepackage{cite}
\usepackage{xcolor}
\usepackage{booktabs}

\title{\LARGE \bf
Population-Aware Physics-Informed Neural Particle Flow for Robust Spacecraft Bayesian Navigation
}

\author{Batu Candan$^{1}$ and Simone Servadio$^{2}$
\thanks{$^{1}$PhD Candidate,
         Iowa State University, Ames, IA, USA,
        }%
\thanks{$^{2}$Assistant Professor,
         Iowa State University, Ames, IA, USA,
        }%
}

\begin{document}

\maketitle 
\thispagestyle{empty} 
\pagestyle{empty}

\begin{abstract}
Spacecraft navigation often requires Bayesian inference from sparse nonlinear measurements that produce curved, multimodal, or geometrically constrained posterior distributions. Physics-informed neural particle flow (PINPF) addresses such problems by learning a deterministic prior-to-posterior transport field from the governing probability evolution equation, but its particle-wise architecture does not explicitly account for the empirical particle population. This paper introduces population-aware PINPF (PA-PINPF), which conditions each particle velocity on a permutation-invariant Deep Sets representation of the complete particle set. PA-PINPF-State summarizes particle positions, whereas PA-PINPF-Feature summarizes the local physics-informed features, including likelihood and score information. Both retain the unsupervised PINPF residual and require neither posterior samples nor labeled transport trajectories. The methods are evaluated over 100 randomized time-difference-of-arrival tasks representative of RF cross-link and ground-station localization and 100 range-measurement tasks with an onboard sensor field-of-view constraint. Results demonstrate that population-level Bayesian features provide useful global information for learned particle transport in nonlinear spacecraft-navigation problems.
\end{abstract}

\section{Introduction}

Bayesian inference is a fundamental component of nonlinear estimation, tracking, localization, and multi-sensor fusion in space applications \cite{sarka}. In these problems, the objective is to represent a posterior distribution over system states after incorporating uncertain measurements. Over the past two decades, a large body of nonlinear filtering methods has been developed to approximate the Bayesian recursive relations when exact posterior updates are not tractable \cite{servadioRev1}. Recent surveys organize these methods into several major families, including point mass filters, particle filters \cite{tuto}, particle flow filters \cite{khan, pffHighd, pffAux, pffInv}, Gaussian sum or Gaussian mixture filters, hybrid Gaussian-mixture/particle filters, and high-order nonlinear filters \cite{servaFlow, servarecursive}. 

PINPF, in contrast, learns a velocity field constrained by the probability evolution equation associated with the Bayesian homotopy from prior to posterior \cite{pinpf}. This combines neural approximation with the structure of particle-flow-based Bayesian transport. However, the standard PINPF velocity model is particle-wise: the velocity of each particle is computed from that particle's state and task information, without explicitly conditioning on the empirical particle population. This can limit the ability of the learned flow to adapt to the actual sample configuration, including its spread, skewness, multimodality, undercoverage, and sample imbalance. 

In this paper, we propose a population-aware physics-informed neural particle flow method for Bayesian posterior transport for spacecraft navigation applications. The key idea is to condition each particle's velocity not only on its own state and task variables, but also on a permutation-invariant representation of the full particle set. This allows the learned velocity field to adapt to the empirical distribution represented by the particles, including its spread, and multimodal structure, while preserving exchangeability with respect to particle ordering. The proposed architecture uses a Deep Sets-style encoder to summarize the particle-state population. It provides this context to the physics-informed neural velocity model, which still receives the standard local PINPF feature vector for each particle \cite{pfbr, deepset, deep2}.

\section{Background}

\subsection{Bayesian Posterior Transport}

Let $x \in \mathbb{R}^{d}$ denote the random state vector, $p_0(x)$ the prior density, and $l(x)=p(y|x)$ the likelihood associated with a measurement $y$. The Bayesian posterior is given by
\begin{equation}
p_1(x) = p(x|y) =
\frac{p_0(x)l(x)}
{\int p_0(x)l(x)dx}.
\end{equation}
Classical particle filters approximate the posterior using weighted particles. Although this representation is flexible, importance weights may become highly concentrated when the likelihood is sharp or when the state dimension increases. This leads to weight degeneracy and often requires resampling, which can reduce particle diversity \cite{chenrev, rev3, pf}.

Particle flow methods provide an alternative approach by transporting particles from the prior distribution to the posterior distribution \cite{mamich, mallik}. A common construction introduces a pseudo-time variable $\lambda \in [0,1]$ and defines the intermediate density,
\begin{equation}
p_{\lambda}(x)
=
\frac{p_0(x)l(x)^{\lambda}}{Z_{\lambda}},
\qquad
Z_{\lambda}
=
\int p_0(x)l(x)^{\lambda}dx.
\end{equation}
At $\lambda=0$, $p_{\lambda}$ is the prior, and at $\lambda=1$, $p_{\lambda}$ is the posterior. Particles are transported according to the ordinary differential equation
\begin{equation}
\frac{d x_{\lambda}}{d\lambda}
=
v_{\lambda}(x_{\lambda}),
\end{equation}
where $v_{\lambda}$ is a velocity field chosen so that the evolving particle distribution follows the density path $p_{\lambda}$.

For deterministic transport, conservation of probability mass is governed by the continuity equation
\begin{equation}
\frac{\partial p_{\lambda}}{\partial \lambda}
=
- \nabla \cdot \left(p_{\lambda} v_{\lambda}\right).
\end{equation}
Combining the log-homotopy definition of $p_{\lambda}$ with the continuity equation yields a physics-based constraint on the velocity field. This constraint forms the basis of physics-informed neural particle flow methods.

\subsection{Physics-Informed Neural Particle Flow}

In the baseline PINPF formulation, the transport velocity is evaluated
independently for each particle. For particle $i$, the local input feature
vector is
\begin{equation}
c_i =
\left[
x_i,\,
\lambda,\,
y,\,
\log l(x_i),\,
\nabla_x \log p_{\lambda}(x_i),\,
\nabla_x \log l(x_i)
\right].
\label{eq:pinpf_features}
\end{equation}
The components of $c_i$ have the following roles:
\begin{itemize}
    \item $x_i \in \mathbb{R}^{d}$ is the current state of particle $i$;
    \item $\lambda \in [0,1]$ is the pseudo-time variable controlling the
    progression from prior to posterior;
    \item $y$ is the current measurement or observation associated with the
    Bayesian update;
    \item $\log l(x_i)=\log p(y|x_i)$ is the log-likelihood evaluated at
    particle $i$;
    \item $\nabla_x \log p_{\lambda}(x_i)$ is the score of the intermediate
    homotopy density and describes the local geometry of the current target
    density;
    \item $\nabla_x \log l(x_i)$ is the likelihood score and indicates the
    local direction in which the measurement increases probability.
\end{itemize}

The score of the intermediate density can be computed from the unnormalized
log-homotopy as
\begin{equation}
\nabla_x \log p_{\lambda}(x_i)
=
\nabla_x \log p_0(x_i)
+
\lambda \nabla_x \log l(x_i),
\end{equation}
because the normalizing constant $Z_{\lambda}$ does not depend on $x$.

The baseline PINPF velocity is then predicted by a multilayer perceptron,
\begin{equation}
v_i = v_{\theta}(c_i),
\end{equation}
where $v_{\theta}$ denotes the particle-wise velocity network with trainable
parameters $\theta$.

For particle $i$ at pseudo-time step $k$, the physics-informed residual is
\begin{equation}
\begin{aligned}
\mathcal{R}_{i,k}
&=
\log l(x_i^k)
-
\frac{1}{N}
\sum_{j=1}^{N}
\log l(x_j^k)
\\
&\quad
-
\left[
-\nabla_x \cdot v_{\theta}(c_i^k)
-
v_{\theta}(c_i^k)^{T}
\nabla_x \log p_{\lambda_k}(x_i^k)
\right].
\end{aligned}
\label{eq:pinpf_residual}
\end{equation}
The first line is the homotopy-driven change in log-density, while the second
line is the change induced by the learned transport field through compression
and advection.

The PINPF training objective is the mean squared residual over all particles
and pseudo-time steps:
\begin{equation}
\mathcal{L}_{\mathrm{PINPF}}
=
\frac{1}{K}
\sum_{k=1}^{K}
\frac{1}{N}
\sum_{i=1}^{N}
\mathcal{R}_{i,k}^{2},
\label{eq:pinpf_loss}
\end{equation}
where $N$ is the number of particles and $K$ is the number of discrete
pseudo-time steps. The factors $1/N$ and $1/K$ convert the sums into averages,
so that the loss scale remains approximately independent of the number of
particles and integration steps.

\section{Population-Aware PINPF}

\subsection{Motivation}

The baseline PINPF formulation learns a local velocity rule of the form $v_i=v_{\theta}(c_i)$. This is computationally efficient and preserves exchangeability across particles. However, the velocity of each particle is computed without direct access to the empirical particle distribution. As a result, two particle populations with different global structures may produce similar local features for a given particle, even though they may require different transport behavior \cite{pinpf}. The current particle population contains information about the empirical approximation of the intermediate density, including spread, skewness, mode coverage, and sample imbalance. This information can be important in nonlinear and multimodal posterior transport problems. For example, in a Gaussian mixture posterior, accurate inference requires not only moving particles toward high-likelihood regions, but also allocating the correct amount of probability mass to each mode. A purely particle-wise velocity model may have limited ability to adapt to such population-level structure \cite{last1, last2}.

\subsection{Permutation-Invariant Population Encoding}

Let
\begin{equation}
C_{\lambda}
=
\{c_1,\ldots,c_N\}
\end{equation}
denote the set of local PINPF feature vectors at pseudo-time $\lambda$. We
construct a fixed-dimensional representation of the particle population using
a Deep Sets encoder. The encoder consists of two multilayer perceptrons:
an element-wise embedding network $\phi$ and a context network $\rho$. The element-wise network $\phi$ is applied independently, with shared
parameters, to every member of the set. The resulting embeddings are averaged
over the population, and the pooled representation is then processed by
$\rho$ to produce the final population context. Mean pooling makes the
representation invariant to particle ordering.

For the state-based variant, each particle state is embedded as
\begin{equation}
r_j^x = \phi_x(x_j),
\end{equation}
and the state-population context is
\begin{equation}
e_{\lambda}^{x}
=
\rho_x
\left(
\frac{1}{N}
\sum_{j=1}^{N}
r_j^x
\right)
=
\rho_x
\left(
\frac{1}{N}
\sum_{j=1}^{N}
\phi_x(x_j)
\right).
\label{eq:state_population_context}
\end{equation}
Here, $\phi_x$ is a shared state-embedding MLP and $\rho_x$ is a context MLP that maps the pooled embedding into a fixed-dimensional vector
$e_{\lambda}^{x}\in\mathbb{R}^{d_c}$. The corresponding transport velocity is
\begin{equation}
v_i^{x}
=
v_{\theta_x}
\left(
[c_i,e_{\lambda}^{x}]
\right),
\label{eq:state_pa_velocity}
\end{equation}
where $[\cdot,\cdot]$ denotes vector concatenation and $v_{\theta_x}$ is the
population-aware velocity MLP. This model, denoted PA-PINPF-State, summarizes
the geometry of the empirical particle cloud, including its location, spread,
orientation, concentration, and mode coverage.

For the feature-based variant, the same Deep Sets construction is applied to
the complete local feature vectors. Each feature vector is first embedded as
\begin{equation}
r_j^c = \phi_c(c_j),
\end{equation}
and the feature-population context is
\begin{equation}
e_{\lambda}^{c}
=
\rho_c
\left(
\frac{1}{N}
\sum_{j=1}^{N}
r_j^c
\right)
=
\rho_c
\left(
\frac{1}{N}
\sum_{j=1}^{N}
\phi_c(c_j)
\right).
\label{eq:feature_population_context}
\end{equation}
Here, $\phi_c$ is a shared feature-embedding MLP and $\rho_c$ is a context MLP
that maps the pooled feature embedding into
$e_{\lambda}^{c}\in\mathbb{R}^{d_c}$.

The feature-population-aware velocity is
\begin{equation}
v_i^{c}
=
v_{\theta_c}
\left(
[c_i,e_{\lambda}^{c}]
\right),
\label{eq:feature_pa_velocity}
\end{equation}
where $v_{\theta_c}$ is the corresponding velocity MLP. This model is denoted
PA-PINPF-Feature.

The two variants therefore use the same overall structure but differ in the
set being summarized:
\begin{align}
\text{PA-PINPF-State:}\qquad
&e_{\lambda}^{x}
=
\rho_x
\left(
\frac{1}{N}
\sum_{j=1}^{N}
\phi_x(x_j)
\right),\\
\text{PA-PINPF-Feature:}\qquad
&e_{\lambda}^{c}
=
\rho_c
\left(
\frac{1}{N}
\sum_{j=1}^{N}
\phi_c(c_j)
\right).
\end{align}
The state encoder represents the empirical distribution in state space,
whereas the feature encoder represents the empirical distribution in the
physics-informed feature space.

\subsection{Physics-Informed Training Objective}

The proposed method retains the same physics-informed learning principle as the baseline PINPF \cite{pinpf}. The population context modifies the neural velocity parameterization, but the learning signal remains the residual of the log-homotopy continuity equation.

For PA-PINPF, the residual becomes
\begin{equation}
\begin{aligned}
\mathcal{R}^{\mathrm{PA}}_{i,k}
&=
\log l(x_i^k)
-
\frac{1}{N}\sum_{j=1}^{N}\log l(x_j^k)
\\
&\quad
-
\left[
-\nabla \cdot v_{\theta}(c_i^k,e^k)
-
v_{\theta}(c_i^k,e^k)^{T}
\nabla \log p_{\lambda_k}(x_i^k)
\right].
\end{aligned}
\label{eq:pa_residual}
\end{equation}
The training objective is
\begin{equation}
\mathcal{L}_{\mathrm{PA}}
=
\frac{1}{K}
\sum_{k=1}^{K}
\frac{1}{N}
\sum_{i=1}^{N}
\left(
\mathcal{R}^{\mathrm{PA}}_{i,k}
\right)^2.
\label{eq:pa_loss}
\end{equation}
Thus, PA-PINPF learns a physics-informed transport field whose velocity depends on both local Bayesian geometry and the global empirical particle distribution.

\section{Experiments}

Both task families are drawn from spacecraft navigation scenarios, and both are evaluated over $100$ Monte Carlo tasks. Task parameters are sampled independently for training and evaluation, so no evaluated configuration is seen during training.

\subsection{Evaluation Metrics}

Posterior accuracy is evaluated using equally sized sets of transported particles and reference posterior samples. Energy distance (ED) measures overall multivariate distributional agreement, while sliced Wasserstein distance (SWD) compares the distributions \cite{ed, swd}. Posterior mean and covariance errors quantify discrepancies in the first two moments using the Euclidean and Frobenius norms, respectively. Lower values indicate better posterior approximation for all metrics.

\subsection{TDOA Fusion Task}

The first family is two-receiver time-difference-of-arrival (TDOA)
localization, representative of RF cross-link and ground-station geometries used for orbit determination. The state is a planar position $x\in\mathbb{R}^{2}$ and the measurement is the difference of ranges to receivers $S_A, S_B$,
\begin{equation}
z = h(x) + \nu,
\label{eq:tdoa_measurement}
\end{equation}
\begin{equation}
h(x) = \|x-S_A\|_2 - \|x-S_B\|_2,
\label{eq:tdoa_likelihood}
\end{equation}
where $\nu\sim\mathcal{N}(0,\sigma^2)$ is Gaussian measurement noise. This measurement model produces a hyperbolic likelihood. When fused with a Gaussian prior, the posterior can become highly non-Gaussian and curved, and in some cases multimodal. 

\subsection{Range Measurement Under a Field-of-View Constraint}
 
The second family adds the constraint that motivates this work. An onboard sensor at the origin returns a range measurement
$z=\|x\|+\nu$, $\nu\sim\mathcal{N}(0,\sigma_z^2)$, and observes only within a cone of half-angle $\theta_{\mathrm{fov}}$ about a boresight $\hat{u}$. Detection is itself informative; it excludes every state the sensor could not have seen. With
\begin{equation}
\psi(x)=\theta_{\mathrm{fov}}
-\arccos\!\left(\hat{u}^{\mathsf{T}}x/\|x\|\right),
\label{eq:psi}
\end{equation}
the posterior support is truncated to $\psi(x)\ge 0$, and the density is not merely small outside the cone but exactly zero. Such a posterior is incompatible with the transport formulation of Sec.~II, whose score $\nabla_x\log p_\lambda$ is undefined wherever $p_\lambda$ vanishes. Following \cite{demars}, its discontinuous indicator is approximated by
\begin{equation}
h_\alpha(x)=\tfrac{1}{2}\!\left[\tanh(\alpha\psi(x))+1\right],
\label{eq:smooth}
\end{equation}
and the relaxed likelihood $l_\alpha=l\,h_\alpha$ is used in the homotopy of Sec.~II without further modification. Because $\psi$ is bounded, $\log h_\alpha$ is finite everywhere, and $p_\lambda$ is smooth and strictly positive for every finite $\alpha$; the relaxation therefore restores the regularity the transport formulation requires rather than circumventing it. 

\section{Results and Discussion}

\begin{table*}[t]
\caption{TDOA results over 100 Monte Carlo Tasks. Values are mean $\pm$ standard deviation.}
\label{tab:tdoa_results}
\centering
\resizebox{\textwidth}{!}{
\begin{tabular}{lccccc}
\toprule
Method
& ED $\downarrow$
& SWD $\downarrow$
& Mean error $\downarrow$
& Covariance error $\downarrow$
& Time/task [s] $\downarrow$\\
\midrule
PINPF
& $0.0533 \pm 0.1431$
& $0.2732 \pm 0.2402$
& $0.3020 \pm 0.3332$
& $0.5065 \pm 0.6135$
& $1.5238$\\

PA-PINPF-State
& $0.0246 \pm 0.0543$
& $0.2103 \pm 0.1651$
& $0.2146 \pm 0.2244$
& $0.3608 \pm 0.4310$
& $1.6670$\\

\textbf{PA-PINPF-Feature}
& $\mathbf{0.0076 \pm 0.0258}$
& $\mathbf{0.1467 \pm 0.0951}$
& $\mathbf{0.0993 \pm 0.1272}$
& $\mathbf{0.2856 \pm 0.2567}$
& $1.6667$\\
\bottomrule
\end{tabular}}
\end{table*}

\begin{table*}[t]
\caption{Constrained Range/FoV results over 100 randomized test tasks.}
\label{tab:fov_results}
\centering
\resizebox{\textwidth}{!}{
\begin{tabular}{lccccccc}
\toprule
Method & med ED $\downarrow$ & mean ED $\downarrow$
& med SWD $\downarrow$ & mean SWD $\downarrow$
& Outside FoV [\%] $\downarrow$ & $\times$ target $\downarrow$
& Viol. angle [deg] $\downarrow$\\
\midrule
PINPF                    & 1.669 & 3.192 & 2.183 & 4.140 & 58.85 & 44.2 & 32.8\\
PA-PINPF-State           & 1.992 & 8.429 & 2.316 & 4.644 & 39.29 & 29.5 & 47.7\\
\textbf{PA-PINPF-Feature}& \textbf{0.351} & \textbf{2.006} & \textbf{0.998} & \textbf{1.891}
                         & \textbf{11.25} & \textbf{8.4} & \textbf{17.7}\\
Incompressible flow      & 1.134 & 1.391 & 2.006 & 2.419 & 23.66 & 17.7 & 90.1\\
SVGD  & 0.398 & 1.350 & 2.150 & 3.562 & 15.92 & 11.9 & 105.4\\
\bottomrule
\end{tabular}}
\end{table*}

\subsection{TDOA Results}

\begin{figure}
    \centering
    \includegraphics[width=0.47\textwidth]{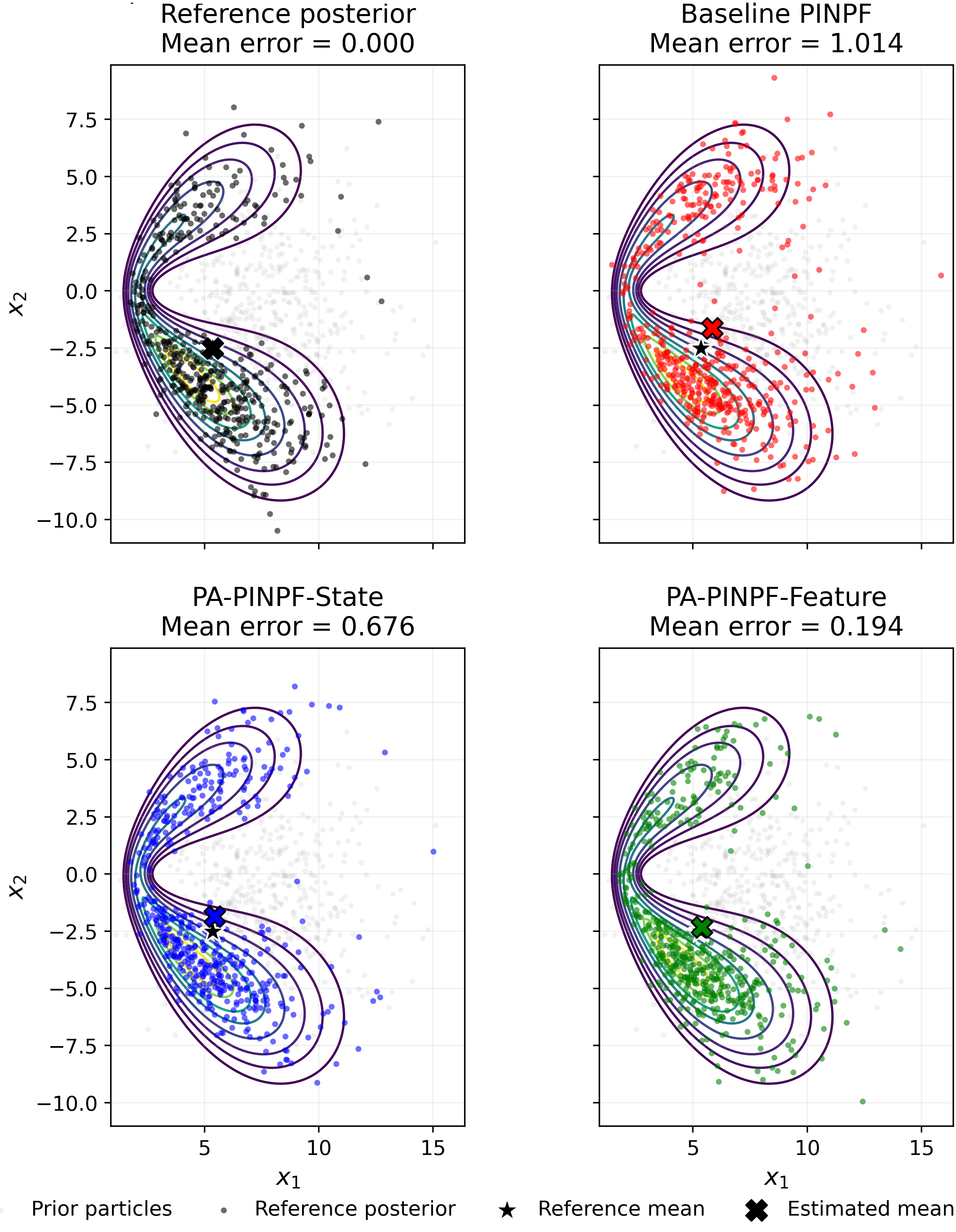}
    \caption{Qualitative comparison on TDOA test task \#87.}
    \label{fig:tdoa_task87}
\end{figure}

Table~\ref{tab:tdoa_results} reports results on 100 Monte Carlo TDOA test tasks. Both population-aware variants improve on the particle-wise baseline. PA-PINPF-State reduces ED from $0.0533$ to $0.0246$ and PA-PINPF-Feature to $0.0076$, an $86\%$ reduction, with posterior mean error falling from $0.3020$ to $0.0993$. These improvements show that feature-population context improves both global distribution matching and posterior moment accuracy on nonlinear TDOA fusion tasks. Figure \ref{fig:tdoa_task87} provides a qualitative comparison for a representative task selected from the test set. 

\subsection{Field-of-View Constraint Results}
 
\begin{figure}
    \centering
    \includegraphics[width=0.50\textwidth]{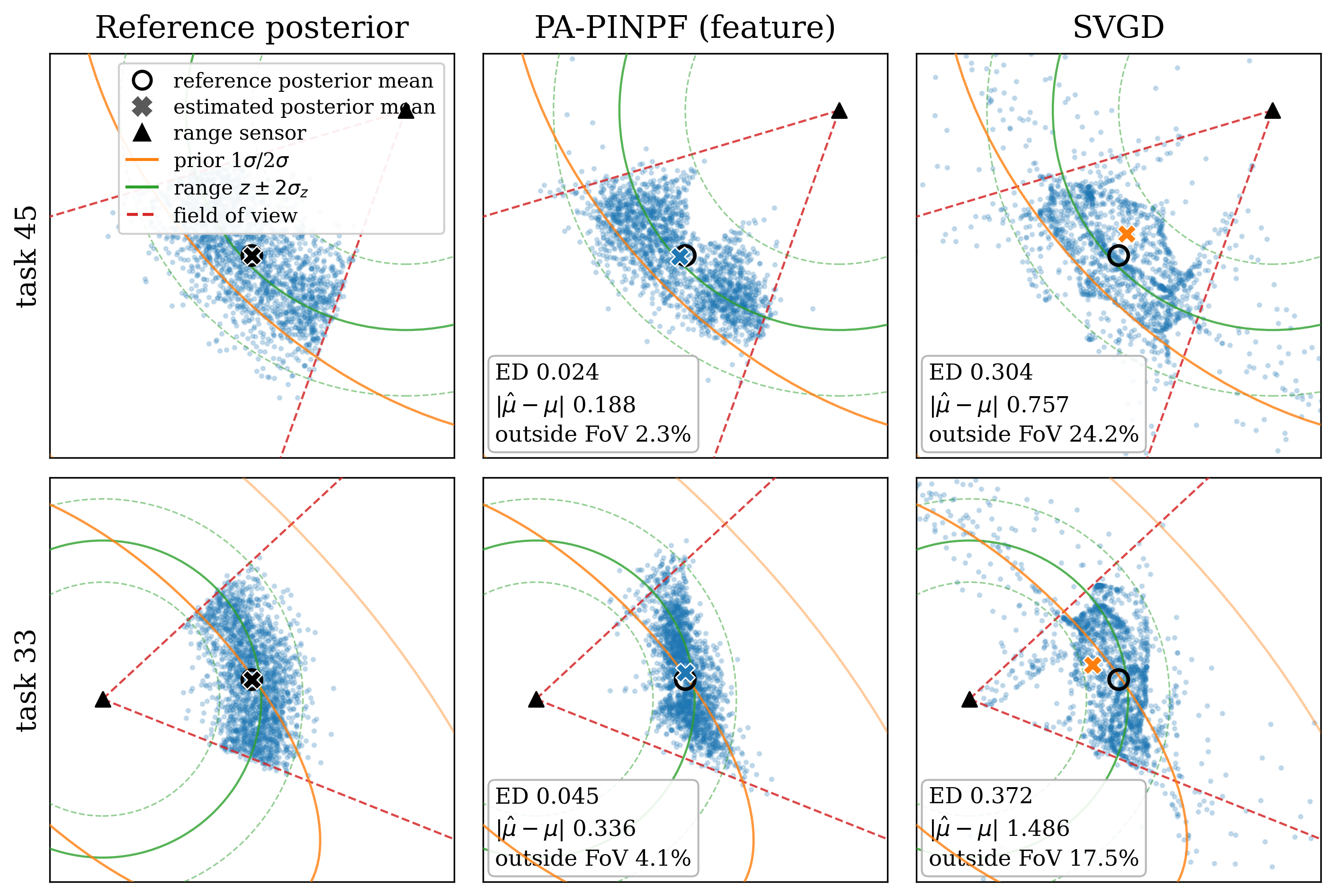}
    \caption{Qualitative comparison on FoV test task \#45 and \#33}
    \label{fig:manu}
\end{figure}

\begin{figure}
    \centering
    \includegraphics[width=0.50\textwidth]{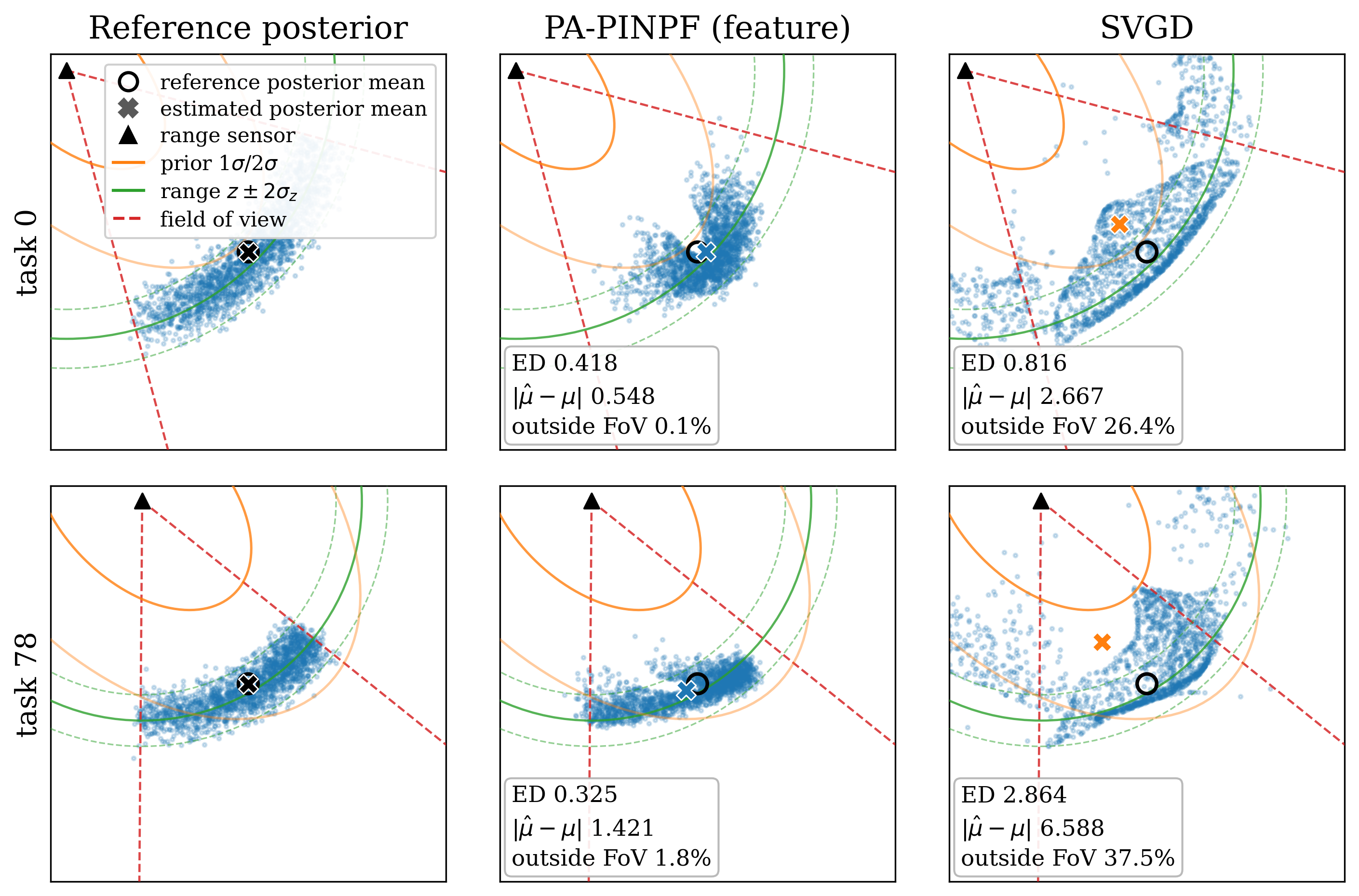}
    \caption{Qualitative comparison on FoV test task \#0 and \#78}
    \label{fig:manu2}
\end{figure}

Table~\ref{tab:fov_results} shows that the constraint changes the character of the comparison. PINPF places $58.9\%$ of its particles outside the field of view and PA-PINPF-State $39.3\%$. PA-PINPF-Feature reaches $11.25\%$, the closest of any transport method. PA-PINPF-Feature instead achieves the best median ED, median SWD, infeasible fraction, and violation angle among the transport methods. It also provides a better overall posterior approximation than SVGD. Although their median ED values are close, SVGD has a $2.2\times$ larger median SWD and a $5.9\times$ larger mean per-task worst angular violation. As illustrated in Figures~\ref{fig:manu} and \ref{fig:manu2}, SVGD approximately recovers the posterior extent but distorts its geometry and places particles far behind the sensor. Although PA-PINPF-Feature does not strictly enforce feasibility, it achieves the lowest violation fraction ($11.25\%$) and angular violation ($17.7^\circ$) among the transport methods, supporting its overall success across both task families.
\section{Conclusion}
This paper introduced population-aware physics-informed neural particle flow for nonlinear Bayesian inference in spacecraft-navigation applications. PA-PINPF preserves the unsupervised probability-evolution residual of PINPF while conditioning each particle velocity on a permutation-invariant representation of the evolving particle population. The state-based encoder represents the geometry of the empirical particle cloud, whereas the feature-based encoder additionally communicates population-level likelihood, score, measurement, and homotopy information. The experiments show that population awareness is most effective when it represents the Bayesian transport geometry rather than particle positions alone. The resulting framework provides a foundation for population-aware inference in RF localization, onboard relative navigation, and constrained spacecraft sensing. Future work will extend the method to sequential filtering, time-varying sensor geometry, and transport mechanisms that more strictly preserve measurement constraints.

\end{document}